\documentclass[10pt,twocolumn,letterpaper]{article}

\usepackage{cvpr}
\usepackage{times}
\usepackage{epsfig}
\usepackage{graphicx}
\usepackage{amsmath}
\usepackage{amssymb}
\usepackage{subfigure}

\usepackage[breaklinks=true,bookmarks=false]{hyperref}

\cvprfinalcopy 

\ifcvprfinal\fi

\begin{document}

\title{DeepPerimeter: Indoor Boundary Estimation from Posed Monocular Sequences}

\author{Ameya Phalak, Zhao Chen, Darvin Yi, Khushi Gupta, Vijay Badrinarayanan  \& Andrew Rabinovich \\
Magic Leap, Inc.\\
Sunnyvale, CA 94089, USA \\
\texttt{\{aphalak, zchen, dayi, kgupta, vbadrinarayanan, arabinovich\}@magicleap.com} \\
}

\maketitle

\begin{abstract}
   We present DeepPerimeter, a deep learning based pipeline for inferring a full indoor perimeter (i.e. exterior boundary map) from a sequence of posed RGB images. Our method relies on robust deep methods for depth estimation and wall segmentation to generate an exterior boundary point cloud, and then uses deep unsupervised clustering to fit wall planes to obtain a final boundary map of the room. We demonstrate that DeepPerimeter results in excellent visual and quantitative performance on the popular ScanNet and FloorNet datasets and works for room shapes of various complexities as well as in multiroom scenarios. We also establish important baselines for future work on indoor perimeter estimation, topics which will become increasingly prevalent as application areas like augmented reality and robotics become more significant. 
\end{abstract}

\section{Introduction}
Understanding the 3D layout of an interior is crucial to understanding the long-range geometry of a space with a myriad of applications in augmented reality, navigation, and general scene understanding. Such layouts can be presented in a variety of ways, from cuboid parameters \cite{zou2018layoutnet} to monocular corner coordinates and their connectivity \cite{lee2017roomnet} to more semantically rich full floor plans \cite{liu2018floornet}. These methods differ in the amount of information they require at input and their assumptions regarding the room geometry (e.g. \cite{liu2018floornet} requires a clean 3D point cloud at input while the other two methods \cite{lee2017roomnet, zou2018layoutnet} require monocular perspective or panorama images). The lack of consistency between this set of related problems reveals a general disagreement over what the standard setting for layout prediction should be for indoor scenes.

\begin{figure}[h]
\centering
\subfigure[]{
\includegraphics[width=0.34\textwidth]{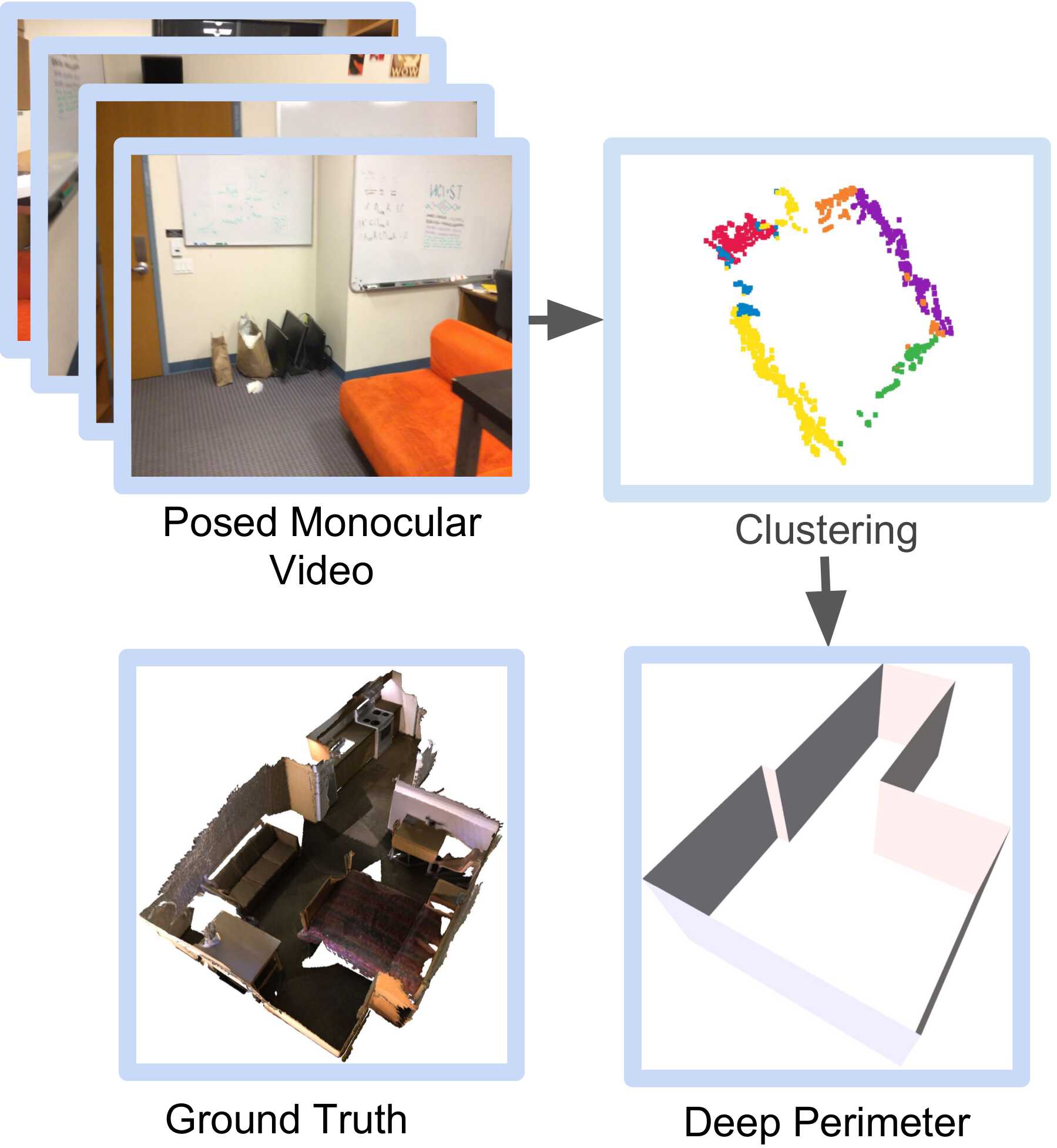}}
    \label{fig:teaser}
    \vspace{-5pt}
    \caption{Our proposed pipeline estimates the indoor perimeter from only a posed monocular video.}
\end{figure}

We thus take this opportunity to ask the question: what types of data are the most readily available in the wild and what type of layout fills the most immediate need for modern vision applications? In terms of sensor data, it is easy to obtain RGB camera and pose data from many modern devices (e.g. smartphones, AR/VR HMDs, etc.). Access to the full video sequence allows us to go beyond the corner and edge estimation that pervade monocular layout estimation and makes possible the estimation of the full \textit{perimeter map} of an interior space\footnote{We use \textit{perimeter} as opposed to \textit{layout} to emphasize that our estimates lie in the world frame and focus on walls. Assuming floors and ceilings are uniplanar we can recover standard monocular layouts (as in \cite{lee2017roomnet}) by reprojecting our perimeter estimates into the camera frame and calculating corner locations from wall intersections.}. Such metric information on the spatial extent and shape of a space is a fundamental invariant of interior spaces (unlike objects within a room which can dynamically shift around), and thus is useful for a variety of downstream 3D applications by enabling area and volume measurements for mobile Augmented Reality platforms such as \cite{arkit}. Our method, DeepPerimeter, is able to leverage current deep methods to precisely infer this perimeter without any hand-crafted enumerated set of the types of possible rooms\cite{lee2017roomnet}. DeepPerimeter is also robust to corner and edge occlusions which are frequent in real world scenes. For this initial study, we only predict the horizontal perimeter (i.e. location of exterior walls), as these contain the vast majority of the structure within a room layout while the floors and ceilings are usually well-approximated by a single plane. See Fig. \ref{fig:teaser} for a basic outline of our pipeline. 

Our pipeline starts with deep depth estimation on the RGB frames of the video sequence. Indeed, one of the most restrictive bottlenecks for general 3D reconstruction applications of deep learning is the accuracy of deep depth estimation models. On cluttered indoor scenes like those in the NYUv2 dataset, such networks still struggle to perform better than 0.5-0.6m of RMS error \cite{laina2016deeper, eigen2015predicting} given monocular inputs. We bypass this performance bottleneck by incorporating multiple views into our depth estimation module by using modern multi-view stereo methods \cite{wang2018mvdepthnet} instead.

We then train a deep segmentation network to isolate the depth predictions corresponding to wall points. These predictions are projected to a 3D point cloud and then clustered through a deep network that is tuned to detect points that belong to the same plane instance. Once point clusters are assigned, standard methods translate the clusters into a full set of planes which form the full perimeter layout. Crucially, by directly clustering wall points, our method is not handicapped when corners are occluded.

Our main contributions are as follows: (1) We propose a deep learning based pipeline for predicting indoor perimeter from a monocular sequence of posed RGB images. Our method is both robust to occluded corners and does not depend on an enumerated set of a priori room shapes. (2) We develop an end-to-end model for clustering point clouds into long-range planar regions using synthetically generated ground truth. (3) We establish important benchmarks for perimeter estimation for future work. 

\section{Related Work}
\textbf{Layout Estimation} Single room layout estimation has been an active research problem since before the deep learning era, although most traditional methods involved vanishing point prediction followed by ad hoc post processing to predict Manhattan layouts \cite{hedau2009recovering, lee2009geometric,gupta2010estimating,schwing2012efficient,fouhey2014people,wang2013discriminative}. After the advent of deep networks for complex vision problems like segmentation, there has been a significant effort to incorporate deep features in layout estimation pipelines \cite{dasgupta2016delay, mallya2015learning, ren2016coarse}. Of special note are efforts by Lee et al. \cite{lee2017roomnet} and \cite{zou2018layoutnet}, who both devised end-to-end approaches to monocular layout estimation. However, it is worth noting that both these methods relied on a priori enumerations of room types, which for full interiors of complex shapes would be prohibitively difficult.

There has also been work in estimation of more complete 3D schematic reconstructions of interiors with both non-deep \cite{okorn2010toward, sui2016layer, turner2015fast} and deep \cite{liu2018floornet} methods. Such work usually assumes a high-quality interior point cloud is available at model input, which is labor-intensive to obtain. Studies on using unlabeled video sequences, which are much more easily obtained, to generate interior layouts are much more rare, which is something we hope the present study delivers.

\textbf{Deep Depth and Segmentation}
One of the successes of deep learning is in its efficacy in the complex visual problems of segmentation and depth estimation. Currently, perhaps the most widely-used deep depth benchmark is the work by Eigen et al. \cite{eigen2015predicting}, who proposed a multitask architecture to predict depth from a monocular image. In general, monocular depth estimation has been a highly active problem \cite{laina2016deeper, kuznietsov2017semi, liu2015deep, ladicky2014pulling, xu2017multi}, but has generally fallen short of the accuracy needed by real applications in the wild. As such, there has been increased interest in moving past purely monocular estimation and incorporating other forms of depth information, whether that be through ordinal depth estimation \cite{chen2016single} depth superresolution and densification \cite{mal2018sparse, chen2018estimating, hui2016depth} or multiview stereo \cite{furukawa2015multi, wang2018mvdepthnet}. In the current work, we make extensive use of multiview stereo methods to take maximum advantage of the availability of a video stream.

Deep learning also brought about a major breakthrough in semantic segmentation, with development of richer and richer segmentation network architectures  \cite{long2015fully, badrinarayanan2017segnet, paszke2016enet, ronneberger2015u, liu2015parsenet} and an increasingly diverse set of target classes. We use a pyramid scene parsing network \cite{zhao2017pyramid} in this work. 

\textbf{Plane Estimation}
Our method for boundary/perimeter estimation relies on accurate clustering of a noisy point cloud into plane instances. Plane estimation has been tackled extensively with traditional methods for point cloud inputs \cite{borrmann20113d, yang2010plane} or stereo inputs \cite{gallup2010piecewise, sinha2009piecewise}. More recently, plane and surface estimation has proven to be a useful regularizer for deep depth estimators \cite{wang2016surge}, and there have been monocular deep methods that have recovered plane parameters in both a supervised \cite{liu2018planenet} and unsupervised \cite{yang2018recovering} manner. To our best knowledge, there has yet to be a deep method for clustering planes in point cloud data, despite the potential applications in layout reconstruction.

\section{Pipeline Overview}
\begin{figure*}[t]
\centering
\includegraphics[width=0.8\textwidth]{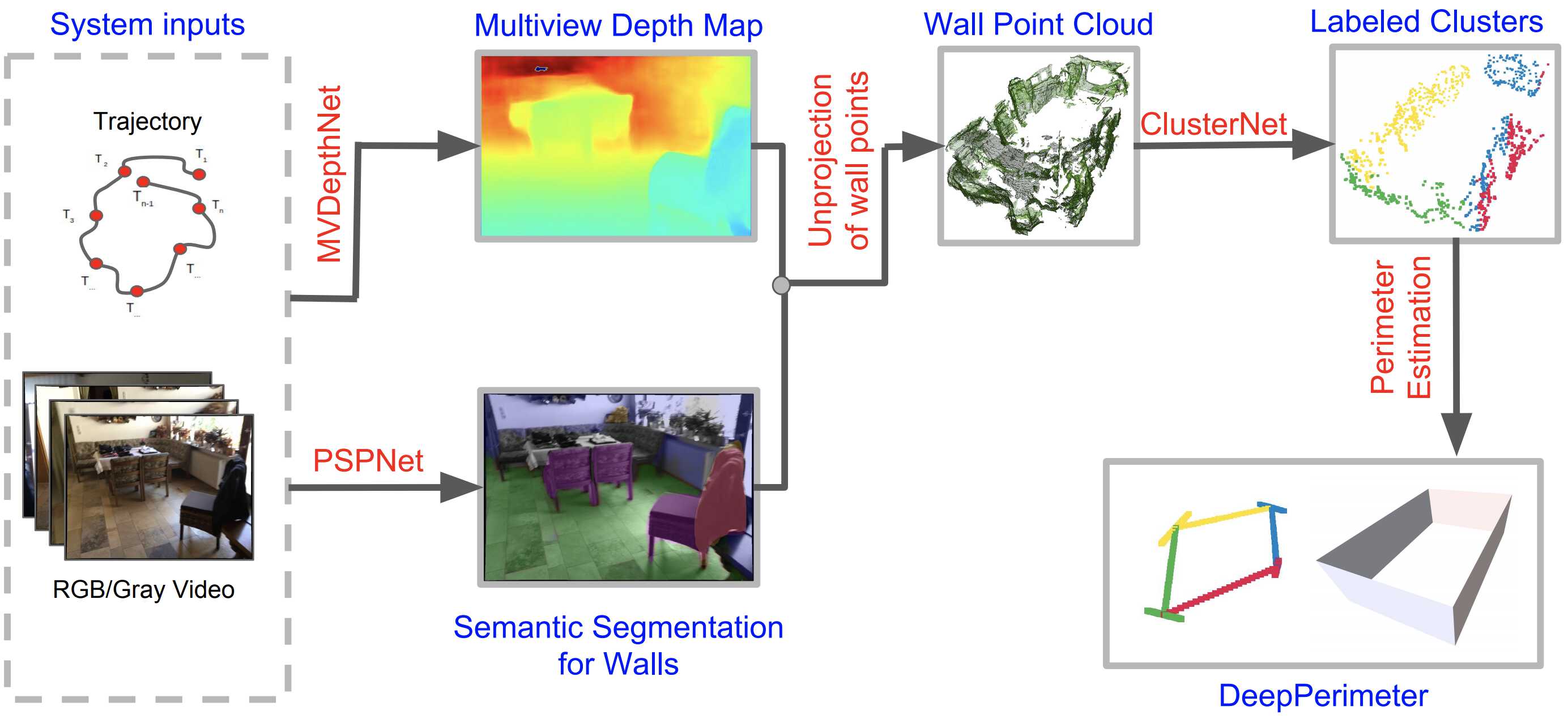}

\caption{Pipeline for perimeter estimation. We begin with a posed monocular sequence of images along with their relative poses. We extract semantic segmentation maps for walls and a dense depth map through a multi-view stereo algorithm. These two outputs are combined through standard unprojection to form a 3D point cloud consisting of wall pixels only. These wall pixels are then colored into wall instance candidates using a deep clustering network and post processed with linear least squares and a shortest path algorithm to form the final perimeter prediction.}
    \label{fig:pipeline}
    \vspace{0pt}
\end{figure*}

Our pipeline consists of two key stages - Deep multi-view depth estimation and segmentation to extract wall point clouds (\ref{sect:mvs}), and deep clustering for detecting plane instances (\ref{sect:clusternet}). The final perimeter is generated in postprocessing by a simple least squares plane fitting (\ref{sect:layout-est}). The full pipeline is illustrated in Fig. \ref{fig:pipeline}. We now describe each of these stages in detail.

\subsection{Extracting Wall Point Clouds}
\label{sect:mvs}
We utilize multiple observations of the same real world scene from various poses to generate a per-frame dense depth map, through the state-of-the art Multiview Depth Estimation network described in \cite{wang2018mvdepthnet}. We then optimize a segmentation algorithm for classifying ceiling, floor, and walls through a standard pyramid scene parsing (PSP) network \cite{zhao2017pyramid} 
with a Resnet-50 \cite{he2016deep} backbone. Details about both the segmentation and depth estimation networks are presented in the Supplementary Materials (Sections \ref{sect:mvs-a} and \ref{sect:sem-seg-a}).


After obtaining a depth map and a wall segmentation mask for each input frame, we generate a unified point cloud using only the depth pixels belonging to the wall class. To do so, we fuse a collection of the masked depth images with known pose trajectory in an implicit surface representation similar to the one described in \cite{curless1996volumetric,newcombe2011kinectfusion} and extract the point cloud by a derivative of the marching cubes \cite{lorensen1987marching} method. The benefit of using an implicit surface representation over simply unprojecting each depth pixel is that it removes redundant points and it averages out the noise over multiple observations leading to a smoother and cleaner set of points as shown in Fig. \ref{fig:pipeline}. 

Finally, to remove internal wall points, we use the concept of $\alpha$-shape as described in \cite{edelsbrunner1983shape} to create a subset of the point cloud that is representative of its concave hull. More details are given in Section \ref{sect:reconstruction-a}. 
\subsection{ClusterNet}
\label{sect:clusternet}
Upon obtaining an $\alpha$-culled, subsampled point cloud representation of the walls in the scene, we proceed to separate the wall instances by a performing a deep clustering of the this point cloud.  

We develop a fully unsupervised technique of clustering unordered point clouds based on planar sections without explicitly computing surface normals or plane parameters during inference. ClusterNet is trained using only the synthetic dataset(see Sect. \ref{sect:syntheticdataset}) and uses a PointNet architecture \cite{qi2016pointnet} with two additional 128$\rightarrow$128 filter convolutional layers right before the prediction head for added capacity. PointNet global features are then used to output a cluster probability for each input point.

In order to generate unique cluster assignments for separate wall instances, we need to be robust to 3D location noise, occlusions and variable point density. Furthermore, the clustering needs to distinguish between parallel planar walls which share the same point normals. We formulate a pairwise loss function that penalizes the network when two points lying on distinct wall instances are assigned the same label. We however do not penalize over-segmentation as cluster merging can be easily accomplished in post-processing.
Take N points $\textbf{x}_i$ with 3D coordinates $P_i$ = $(x_i,y_i,z_i)$, point normals $\textbf{x}^{(n)}_i$ = $(x^{(n)}_i,y^{(n)}_i,z^{(n)}_i)$ and predicted cluster probability vector $P(\textbf{x})$ = $(p^{(\textbf{x})}, ..., p_{k+1}^{(\textbf{x})})$. The $(k+1)^{th}$ class is reserved for points which cannot be confidently placed onto any plane instance to allow the network the ability to filter out noisy points.

The clustering loss $L^{cluster}$ is given as
\begin{equation}
L^{cluster} = \sum_{i,j>i}\mathcal{P}(\textbf{x}_i,\textbf{x}_j)\mathcal{D}(\textbf{x}_i,\textbf{x}_j)
\end{equation}
where the sum is taken over all discrete pairs of points, and
\begin{equation}
\mathcal{P}(\textbf{x}_i, \textbf{x}_j) = \sum_{a=1}^k p^{(\textbf{x}_i)}_a  p^{(\textbf{x}_j)}_a
\end{equation} is the pairwise dot product of the first k elements of the predicted cluster probabilities, and
\begin{equation}
\mathcal{D}(\textbf{x}_i, \textbf{x}_j) = (\textbf{x}_i - \textbf{x}_j) \cdot \textbf{x}^{(n)}_i +  (\textbf{x}_j - \textbf{x}_i) \cdot \textbf{x}^{(n)}_j.
\end{equation}

The term $\mathcal{D}(\textbf{x}_i, \textbf{x}_j)$ has a high value when $\textbf{x}_i, \textbf{x}_j$ lie on different planes (including parallel planes) and a low value when $\textbf{x}_i, \textbf{x}_j$ lie on the same plane. Further, if $\textbf{x}_i, \textbf{x}_j$ have a similar cluster assignment probability vector, $\mathcal{P}(\textbf{x}_i, \textbf{x}_j)$ will be high. We thus penalize when two pairs belong to distinct walls but have a similar cluster assignment probability.

To prevent a trivial solution where all points are assigned to the $(k+1)^{th}$ cluster, we use the regularization loss $L^{reg}$ introduced in \cite{yang2018recovering} as follows:

\begin{equation}
L^{reg} = \sum_{\textbf{x}}\sum_{a=1}^k -1 \cdot \log(p^{(\textbf{x})}_a)
\end{equation}

and thus we penalize whenever the probability a point belongs to \textit{any} plane is small. The total loss is then $L^{cluster} + \beta L^{reg}$ where we set $\beta$ to 1.0 in our experiments.

\subsection{Perimeter estimation}
\label{sect:layout-est}

\begin{figure}[h]
\centering
\subfigure{
\includegraphics[width=0.5\textwidth]{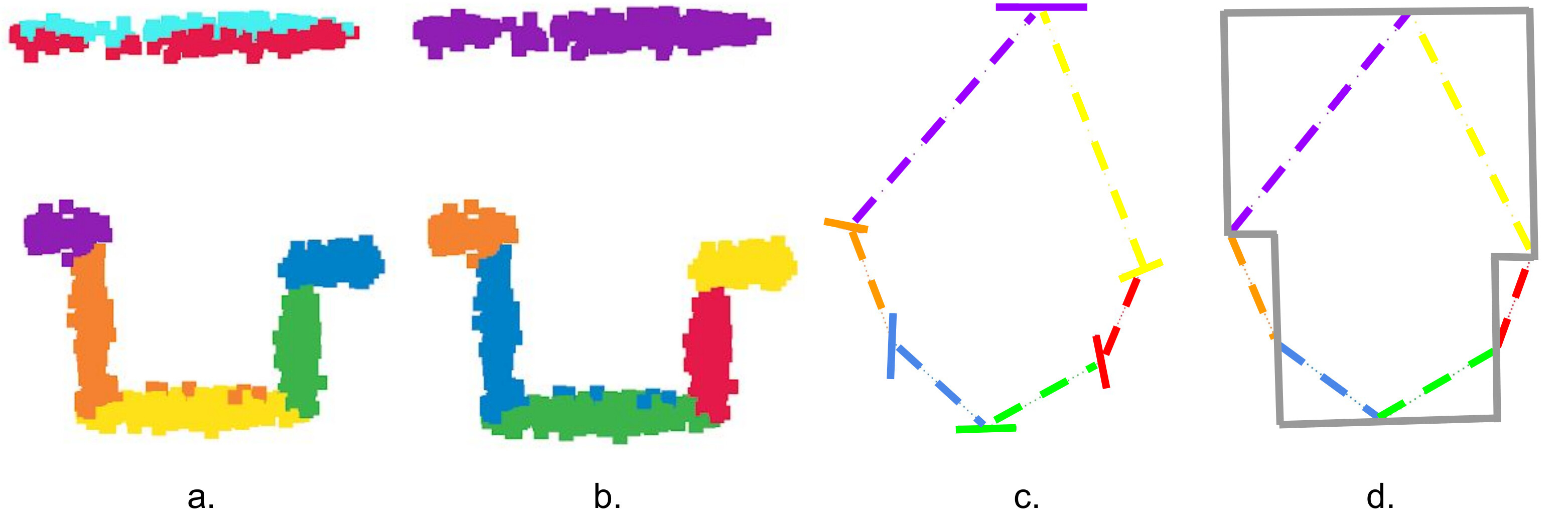}}
\caption{Stages of Perimeter estimation from a clustered point cloud : (a) - Raw ClusterNet outputs with possible oversegmentation. (b) - Compact clustering after merging duplicate clusters. (c) - Estimated line parameters and inter-cluster connectivity. (d) - Extended lines intersect to form corners. Orthogonality and intersections are  forced on connected parallel lines to generate a closed perimeter}
    \label{fig:tsp}
\end{figure}

Following the assumption that all walls are parallel to the Z-axis, we project all clustered 3d wall points to the XY plane to generate a top-down view of the point cloud (see Fig. \ref{fig:tsp}a). We estimate 2D line parameters for every cluster using linear least squares. To remove duplicate wall predictions, clusters having line parameters with a relative normal deviation of $<\theta_{merge}=30^{\circ}$ and an inter-cluster point-to-line error of $<e_{merge}=0.3m$ are assigned the same label as seen in Fig. \ref{fig:tsp}b. To establish connectivity among the clusters, we find a closed shortest path along all cluster medians. The solution is calculated using the algorithm described in \cite{croes1958method} based on the traveling salesman problem and the 2D line parameters are re-estimated as shown in Fig. \ref{fig:tsp}c. 

Resultant lines are then snapped to the nearest orthogonal axis and trivially extended to intersect as shown in  Fig. \ref{fig:tsp}d. The intersection point of two neighboring wall segments is defined as a corner. When major occlusions occur and result in two connected parallel segments, we extend an endpoint of one of the segments in an orthogonal direction so as to force an intersection. Note that such occurrences are rare, and certainly much rarer than the corner occlusion which cause issues in many layout estimation pipelines that operate through direct corner prediction.

\section{Datasets}
In this section we describe the datasets used for training the various networks, including the synthetically generated dataset for training our deep clustering model. 

\subsection{Public Datasets}
We focus on the following publically available datasets to evaluate our various models.
\begin{enumerate}
    \item The ScanNet dataset \cite{dai2017scannet} to train our depth network, evaluate our estimated perimeters, and evaluate our depth and segmentation modules.
    \item The ADE20K \cite{zhou2018semantic} and SunRGBD \cite{song2015sun} datasets for training our segmentation network.
    \item The FloorNet dataset \cite{liu2018floornet} for additional evaluation of our inference of perimeters from unclustered point clouds.

\end{enumerate}

We discuss these datasets in more detail in the Supplementary Materials (Section \ref{sect:pub_data_a}).

\subsection{Synthetic Dataset for ClusterNet}

\begin{figure}[h]
    \subfigure{
    \includegraphics[width=0.5\textwidth]{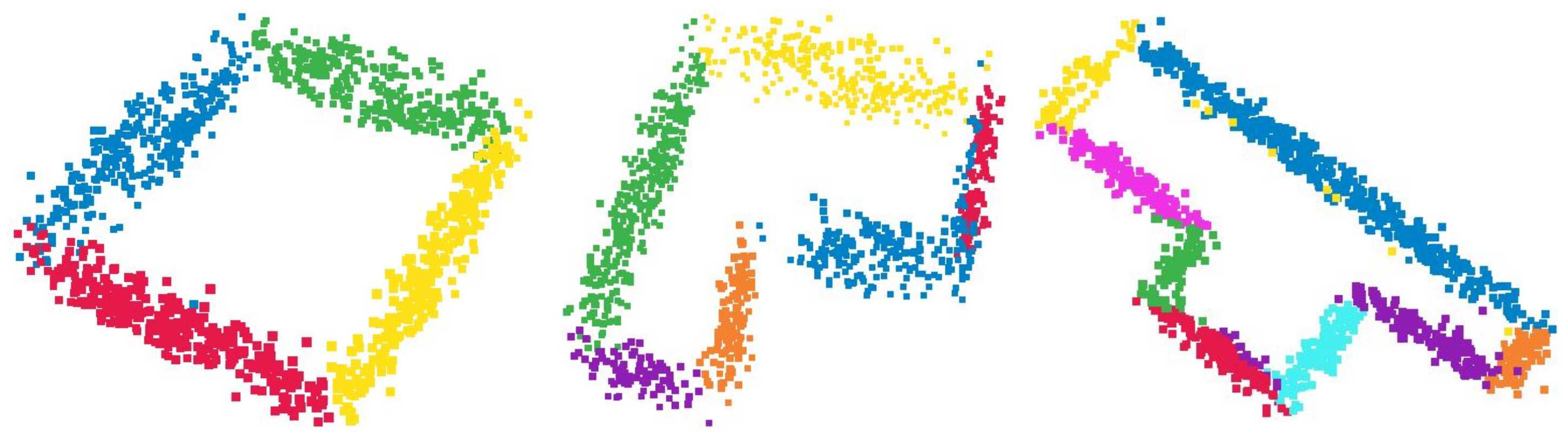}}
\caption{ClusterNet output labels for our synthetic dataset} 
    \label{fig:syntheticdata}
    \vspace{-5pt}
\end{figure}

\label{sect:syntheticdataset}
We build a fully synthetic dataset(see Fig. \ref{fig:syntheticdata}) along with normal labels, starting from a room perimeter skeleton randomly sampled from various shapes (rectangle, L-shaped, T-shaped, or U-shaped). Lengths and angular orientation of each edge and the height of the room are uniformly sampled. Gaussian noise is added and we also included random deletion of points within cylindrical areas to mimic missing points that commonly occur in point cloud measurements.


\section{Results}

We now detail experiments on perimeter estimation using our pipeline. We focus on performance of our clustering and perimeter estimation; we detail experiments on depth estimation and wall segmentation in the Supplementary Materials. All models were run on a single GTX 1080Ti GPU. 

\subsection{ClusterNet Results}

We use the synthetic dataset described in Sect. \ref{sect:syntheticdataset} to train the ClusterNet in a fully unsupervised manner. We generate $10,000$ training examples and $1000$ validation examples. Although the training makes use of synthetic point normals for computing the loss, they are not an input to the network and thus are not necessary during inference. By avoiding explicit computation of normals for each point, we are able to make the clustering efficient. A CPU implementation of the point cloud generation module (see Sect. \ref{sect:reconstruction}) takes around ~8-10s for a sequence of ~300 depth frames, whereas an unoptimized CPU version of the $\alpha$-culling takes $~5s$ for a scene.  A forward pass through the network for a point cloud of size $N=1280$ takes $~5ms$. Labels predicted by ClusterNet on some samples of the synthetic dataset are shown in Fig. \ref{fig:syntheticdata}.

\subsection{Perimeter Estimation Results}

\begin{figure}[h]
    \centering
    \subfigure{
    }\includegraphics[width=0.45\textwidth]{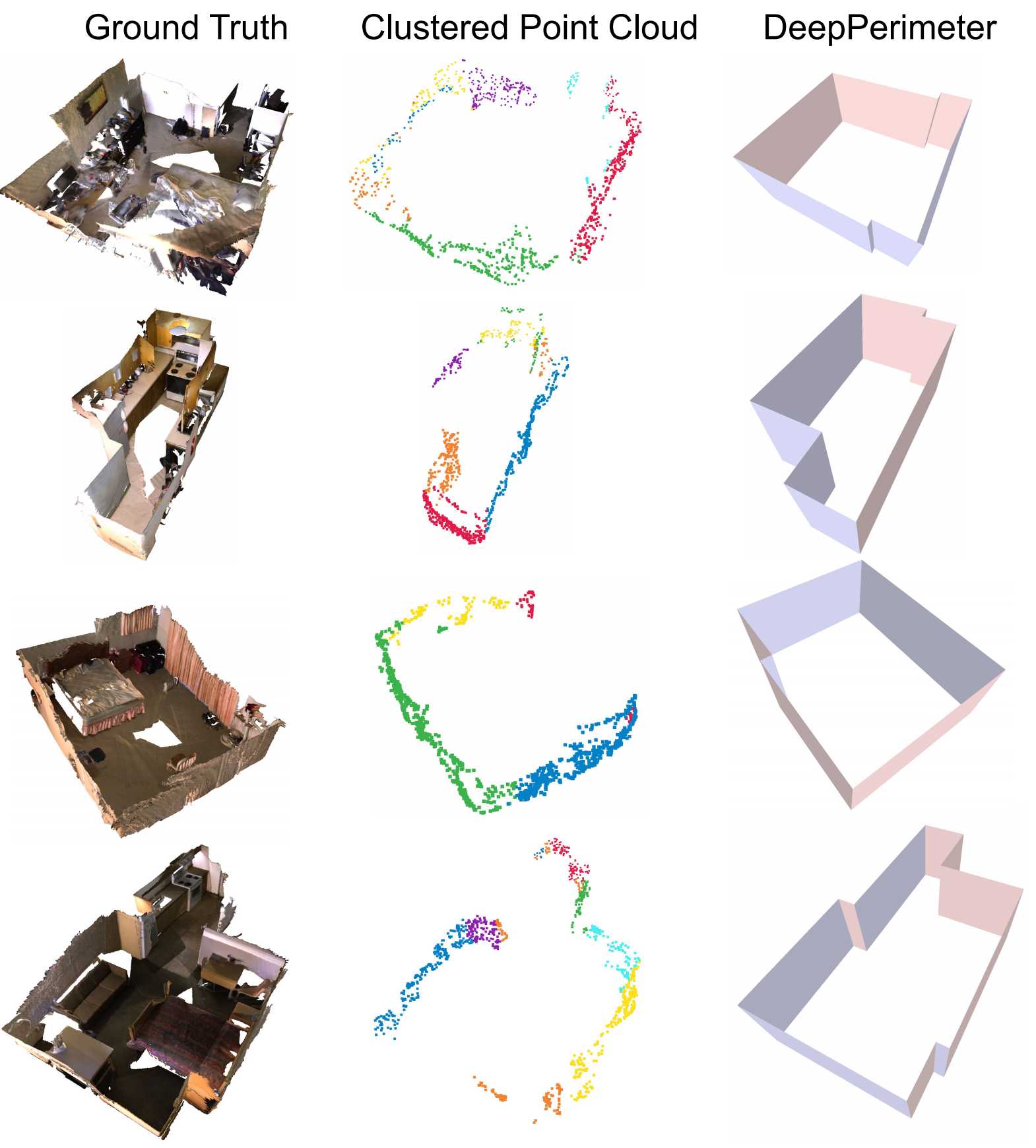}
\caption{ScanNet results. The first column shows the textured ground truth point cloud. The second column displays clusters on the inferred point clouds, and the third column shows a popup visualization of the final perimeter estimate. } 
    \label{fig:resultsscannet}
    \vspace{-5pt}
\end{figure}

\begin{figure}[h]
    \centering
    \subfigure{
    \includegraphics[width=0.45\textwidth]{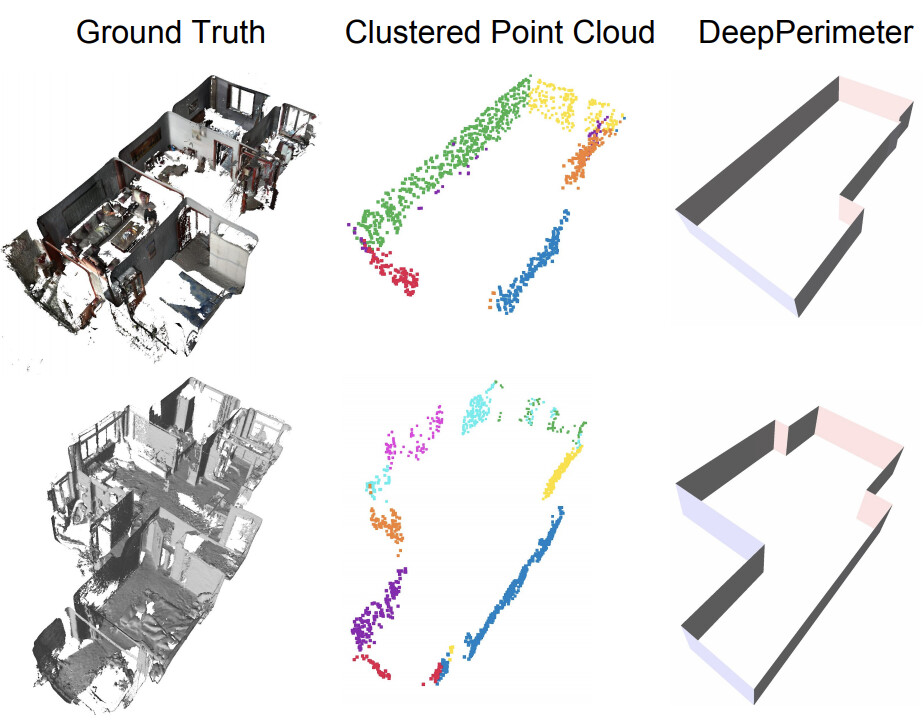}
    }
\caption{Floornet results. The first column shows the ground truth point cloud. The second column displays clusters on the filtered point clouds, and the third column shows a popup visualization of the final perimeter estimate.} 
    \label{fig:resultsfloornet}
    \vspace{-5pt}
\end{figure}

\begin{figure}[h]
    \centering
    \subfigure{
    \includegraphics[width=0.5\textwidth]{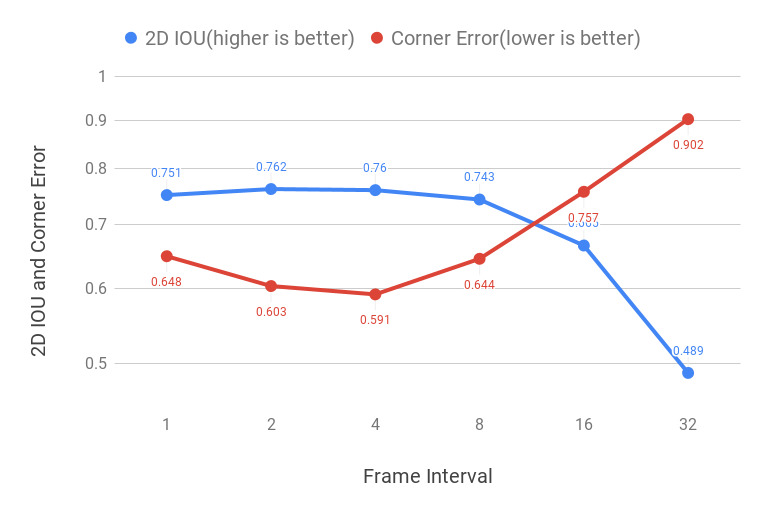}}
\caption{Effect of sampling number of depth frames by a factor of 1, 2, 4, 8, 16, 32 to generate the input points for ClusterNet} 
    \label{fig:frameablation}
\end{figure}

We present visualizations of the final estimated perimeters in Fig. \ref{fig:resultsscannet} for ScanNet and Fig. \ref{fig:resultsfloornet} for FloorNet. In both cases, our pipeline produces perimeter estimates that fit to the room shape well for very cluttered indoor scenes and regardless of the complexity of the shape. We tabulate the quantitative performance of our method in Fig. \ref{fig:frameablation} and Fig. \ref{table:ablation}. Our metrics include the standard IoU of our predicted perimeter with the ground truth, the relative absolute distance between predicted corners and the nearest ground truth corner, and the percentage of spurious corners predicted by our inferred perimeters. We achieve a 0.760 IoU score on the ScanNet dataset and 0.591m absolute error on our predicted corners (which we inferred from the intersection points of planes), which is a strong result especially as we never learn to directly predict corner positions. 

We also find that our method has good performance on FloorNet, achieving a 2D IoU of 0.796 and average relative corner error of 1.279m (See Fig. \ref{fig:resultsfloornet}. The higher IoU is to be expected as we are only able to evaluate the latter half of our pipeline on FloorNet's high-quality LiDAR scans due to the unavailability of poses for its RGB sequences. The higher corner error is also to be expected as FloorNet is predominantly complex multi-room scenes. 

\begin{figure}[h]
    \centering
    \subfigure{
    \includegraphics[width=0.4\textwidth]{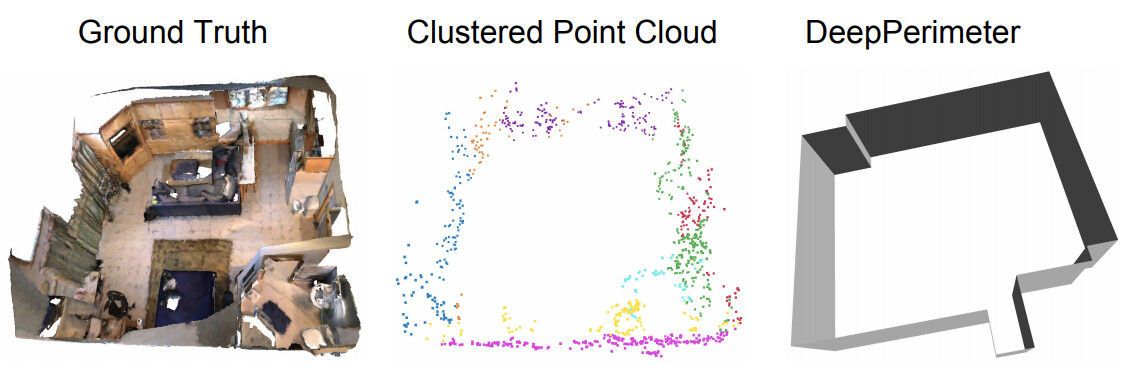}}
\caption{When $\alpha$-culling is disabled, non-perimeter(internal) walls points are not eliminated, leading to erroneous layouts} 
    \label{fig:noalpha}
    \vspace{-5pt}
\end{figure}

We conduct ablation studies to observe the effect of various hyperparameters within our pipeline. We see from Fig. \ref{fig:frameablation} that taking fewer frames from the video sequence for our point cloud reconstruction does not damage the model performance up to a frame interval of 8, which is a potential source of time savings. In Table \ref{table:ablation} we see that including segmentation significantly reduces the corner error and 2D IoU, while the internal wall ($\alpha$) culling significantly reduces the number of spurious corners (see Fig. \ref{fig:noalpha}). We do observe that corner error decreases without segmentation of $\alpha$ culling, but only because of the large increase of spurious corners. This effect artifically deflates corner error as corner error measures the distances between ground truth corners and the \textit{closest} predicted corner.

\begin{table}[ht]
\resizebox{8.5cm}{!}{
\begin{tabular}{|c|c|c|c|}
    \hline
    Technique & 2D IoU & Corner Error(m) & Spurious Corners\\
     & &  & $(\%/100)$\\
    \hline
    Using both & \textbf{0.760} & \textbf{0.591} & \textbf{0.359}\\
    \hline
    W/O Segmentation & 0.762 & 0.611 & 0.480\\
    \hline
    W/O $\alpha$ Culling & 0.778 & 0.491 & 0.856\\
    \hline
    W/O Segmentation or $\alpha$ culling &  0.759 & 0.550 & 1.071\\
    \hline
\end{tabular}}
 \vspace{5pt}
\caption {Effect of removing the $\alpha$-culling and semantic segmentation modules. Spurious corners are the number of extra corners found by DeepPerimeter (expressed as a percentage of the number of ground truth corners).}
\label{table:ablation}
\vspace{-5pt}
\end{table}

\subsection{Limitations}

\begin{figure}[h]
    \centering
     \subfigure{
    \includegraphics[width=0.4\textwidth]{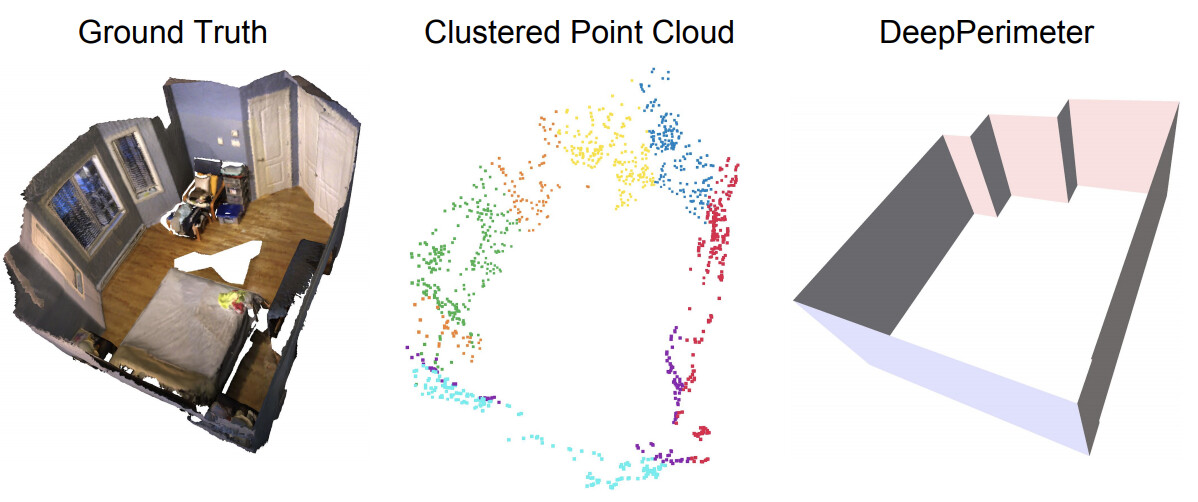}}
\caption{Limitations: Manhattan type of layout is generated when true walls are non-orthogonal} 
    \label{fig:failure1}
    \vspace{-10pt}
\end{figure}

In our experiments we follow the Manhattan assumption which leads to the generated perimeter being sub-optimal for cases where the true walls are non-orthogonal to each other as seen in Fig. \ref{fig:failure1}. A simple solution is to allow for walls which lie at common degree angles (e.g. 45 degrees). 

A limitation introduced by using the plane merging hyperparameter $e_{merge}$ in Sect. \ref{sect:layout-est} is that in cases where walls are separated by an inter-planar distance of less that $e_{merge}$, they will be incorrectly assumed to belong to the same cluster and hence predicted as a single wall. The necessity of $e_{merge}$ comes from noise in the point cloud, which can be alleviated with more robust depth estimation modules. 

\section{Discussion}

Crucially, our pipeline makes very mild assumptions about the input data. We do not assume that high-quality, expensive precomputed point clouds (e.g. from LiDAR) are available at input\cite{liu2018floornet}. Recent smartphones have capabilities to capture pose along with RGB video sequences. Thus, our method can be deployed in common scenarios and is available to standard users without highly specialized equipment. Future work can further alleviate these requirements, e.g. we can infer pose from a video sequence in an unsupervised fashion, as in \cite{yin2018geonet}. 

Our pipeline is also highly modular; as deep learning technologies evolve, we can easily replace, for instance, our segmentation module or depth estimation module with more efficient/accurate variants while retaining the rest of the pipeline. We can also increase efficiency by solving depth and segmentation in a multitask fashion.  

DeepPerimeter represents one of the first attempts to bridge the gap between what data is readily available for everyday users and what invariant 3D information on an interior scene is immediately useful for downstream applications. Now that we can robustly estimate a perimeter from a sequence of posed images, we can potentially use that prediction as scaffolding for populating the interior with other objects of semantic interest (e.g. chairs, tables, various surfaces), eventually creating CAD-like full interior maps from just a posed video sequence. Such model-based maps could form the basis for many diverse use cases, such as robot navigation or augmented reality. Although full, detailed semantic parsing of a scene is beyond the scope of the current study, we believe that the perimeter of the interior is an essential first anchor for research in that direction.

\section{Conclusion}

DeepPerimeter, our proposed deep learning based pipeline for metric-scale estimation of the interior perimeter, is an important step towards widely-applicable large-scale scene understanding of an interior space. It leverages existing deep methods while developing new deep modules to take posed RGB inputs that are widely available and infers from them a piece-wise planar representation of a large space. In contrast with existing room or floor plan layout estimation methods, our pipeline doesn't require expensive LiDAR point clouds or assumptions about possible room shapes. Our work furthermore serves as a natural complement to work in object fitting for the modeling of parametric large-scale scenes by providing an intermediate 3D representation of the room that is geometrically meaningful.
\\ 

%


We would like to acknowledge and thank Chen Liu for help with initial ideas and useful discussion.
\newpage

{\small
\bibliographystyle{ieee}
\bibliography{main}
}

\newpage

\section{Supplementary Materials}
\label{sect:appendix}
\subsection{Extracting Wall Point Clouds}
\label{sect:mvs-a}
We utilize multiple observations of the same real world scene from various poses to generate a per-frame disparity map, which can then be inverted to generate a per-frame dense depth map. We employ a state-of-the art Multiview Depth Estimation network \cite{wang2018mvdepthnet}. The input to this network is an RGB image $I_i$ and a cost volume $C_i^{(z)}$ which we construct at different fixed depths $z$ by calculating pixel-wise absolute intensity differences between $I^{(j)}_i$ and and a neighboring frame $I_j$, where $I^{(j)}_i$ denotes $I_i$ after it has been projected into the frame of $I_j$. For a fixed depth $z$, to find the pixel location $\tilde{\textbf{u}}^{(z)}$ in frame $j$ from pixel location $\textbf{u}$ in frame $i$, we use the (relative-to-world) pose of the reference frame $T_i$, pose of the neighboring frame $T_j$, and shared camera intrinsic transformation $\pi$:

\begin{equation}
    [\tilde{\textbf{u}}^{(z)}, 1]^T\propto
    \pi T_j T_i^{-1}(\textbf{z}*\pi^{-1}[\textbf{u}, 1]^T) 
\end{equation}

\noindent Here $\textbf{z} = [z,z,z,1]^T$ and $\pi$ removes the homogeneous coordinate and left multiplies by the camera intrinsic matrix $K$. We then recover $\tilde{\textbf{u}}$ by normalizing the result to its homogeneous coordinate. By varying $z$ between $z_{min}$ and $z_{max}$, the value of the 3D cost volume for $I_i$ at location $\textbf{u}^{(z)}$ can be calculated as

\begin{equation}
    C_i^{(z)}(u,v) =
    | I_i(\textbf{u}) - I_j(\tilde{\textbf{u}}^{(z)})|.
\end{equation}

To generate a cost volume using multiple neighboring frames at each depth, we simply average the all cost volumes between the current frame and its neighbors. The final cost volume $C_i$ is a concatenation of all cost volumes calculated at different $z$'s; in our implementation we use 64 depth values (see Sect. \ref{sect:mvsdepth_exp}). This averaged cost volume, along with $I_i$ is fed into the network to generate the per pixel depth corresponding to $I_i$.

\subsection{Semantic Segmentation for Walls}
\label{sect:sem-seg-a}

Our primary goal for our semantic segmentation network is to filter for wall locations within the scene, as they are the only points that belong to the interior perimeter. To optimize a segmentation algorithm for classifying ceiling, floor, and walls, we trained a standard pyramid scene parsing (PSP) network \cite{zhao2017pyramid} 
with a Resnet-50 \cite{he2016deep} backbone. We ignored the auxilary classification loss mentioned in the original PSPNet implementation \cite{zhao2017pyramid} and only used cross-entropy loss while finetuning, but modified the datasets to focus on planar surfaces (e.g. walls). See Sect. \ref{sect:wallsegdata} for more details on segmentation data processing. 


\subsection{Multiview Stereo Depth Estimation}
\label{sect:reconstruction}
After obtaining a depth map and a wall segmentation mask for each input frame, we generate a unified point cloud using only the depth pixels belonging to the wall class. To do so, we fuse a collection of the masked depth images with known pose trajectory in an implicit surface representation similar to the one described in \cite{curless1996volumetric,newcombe2011kinectfusion} and extract the point cloud by a derivative of the marching cubes \cite{lorensen1987marching} method. The benefit of using an implicit surface representation over simply unprojecting each depth pixel is that it removes redundant points and it averages out the noise over multiple observations leading to a smoother and cleaner set of points as shown in Fig. \ref{fig:pipeline} 

To remove internal wall points, we use the concept of $\alpha$-shape described in \cite{edelsbrunner1983shape} to create a subset of the point cloud that is representative of its concave hull. First, a delaunay triangulation \cite{chew1989constrained} of the points projected on the $XY$ plane is computed and edges of triangles with a circumcircle of radius $r_{circle} > 1.0 / \alpha$ are discarded. Then, of the remaining triangles, only the boundary points and edges(with a degree of 0, 1 or 2) are retained and interpolated to generate a dense contour. Following this, any point that does not lie within a distance of $d_{cull}$ of the contour is discarded. In our experiments, we set $\alpha=0.5$  and $d_{cull}=0.5m$. The $\alpha$-filtered point cloud is further sub-sampled to exactly $N$ points. We find that complex perimeters can be expressed using $N=1280$, which leads to an efficient representation.

\subsection{Point Cloud Reconstruction with $\alpha$ culling}
\label{sect:reconstruction-a}
To remove internal wall points, we use the concept of $\alpha$-shape as described in \cite{edelsbrunner1983shape} to create a subset of the point cloud that is representative of its concave hull. More details are given in Section \ref{sect:reconstruction-a}. First, a delaunay triangulation \cite{chew1989constrained} of the points projected on the $XY$ plane is computed and edges of triangles with a circumcircle of radius $r_{circle} > 1.0 / \alpha$ are discarded. Then, of the remaining triangles, only the boundary points and edges(with a degree of 0, 1 or 2) are retained and interpolated to generate a dense contour. Following this, any point that does not lie within a distance of $d_{cull}$ of the contour is discarded. In our experiments, we set $\alpha=0.5$  and $d_{cull}=0.5m$. The $\alpha$-filtered point cloud is further sub-sampled to exactly $N$ points. We find that complex perimeters can be expressed using $N=1280$, which leads to an efficient representation.

\subsection{Deep multiview depth estimation}\label{sect:mvsdepth_exp}

\begin{table}[h]
\centering
\begin{tabular}{|c|c|c|}
    \hline
    Method & Rel  & RMSE(m) \\
    \hline
    FPNNet(monocular) \cite{lin2017feature}& 0.150 & 0.325\\
    \hline 
    Ours(1 neighbor) & 0.129 & 0.262\\
    \hline
    Ours(2 neighbors) & 0.115 & 0.237\\
    \hline
    Ours(All neighbors) & \textbf{0.108} &  \textbf{0.225}\\
    \hline
\end{tabular}
 \vspace{5pt}
\caption {Evaluation of the multi-view depth estimation module on the ScanNet test scenes(0.5-5m range). Lower is better for all values.}
\label{table:mvsinference}
\end{table}
We start with the open-source pretrained model from \cite{wang2018mvdepthnet}, and finetune it on the ScanNet dataset. As in \cite{wang2018mvdepthnet}, we use 64 values of fixed depth in the cost volume which are uniformly interspersed in the inverse depth space between $z_{min}$ and $z_{max}$ of the dataset. For calcluating the cost volume, we use all images within a camera translation of $0.3m$ and a camera rotation within $15$ degrees, which we found leads to a lower training and validation error. We evaluate using this model on ScanNet using cost volumes constructed with a varying number of frames (see Table \ref{table:mvsinference}). Constructing the cost volume for a single pair takes around $~900ms$; however, this operation is highly parallelizable for multi-pair cost volumes. A forward pass through the network for an image of size $320\times256$, after the cost volume has been constructed takes $~8ms$.

We see from the results in Table \ref{table:mvsinference} that multiview stereo (MVS) outperforms monocular state-of-the-art methods by a significant margin, with more than a 30\% improvement in RMS error. The MVS error improves as more frames are included in the cost volume information, with diminishing returns. In our final model we chose to use all neighboring frames within a baseline of $0.3$m and $15$ degrees, resulting in $3$ and $4$ neighbors for the test sequences.

\begin{figure}[h]
\subfigure{
\includegraphics[width=0.5\textwidth]{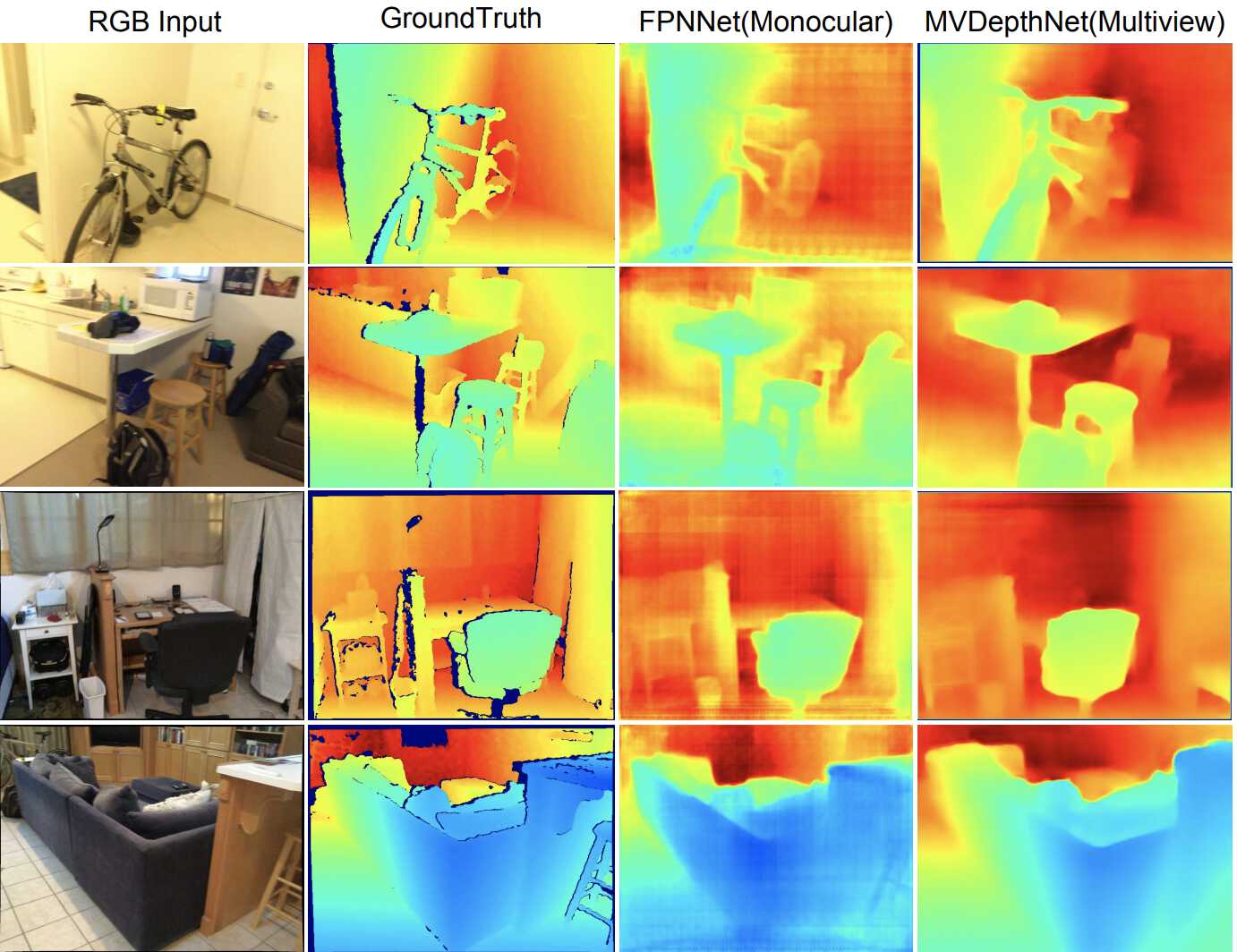}}
    \label{fig:mvsresults}
    \vspace{0pt}
\caption{Depth estimation results. It can be seen that the depth maps generated using the multi-view network are able to resolve finer details and do not suffer from checkerboard artifacts when compared to the monocular network. The inferred depth maps are also more complete and do not have missing values when compared to the ground truth depth maps} 
    \label{fig:resultsmvs}
    \vspace{-5pt}
\end{figure}

\subsection{Wall Segmentation}\label{sect:segresults}

\begin{figure}[h]
\subfigure{
\includegraphics[width=0.5\textwidth]{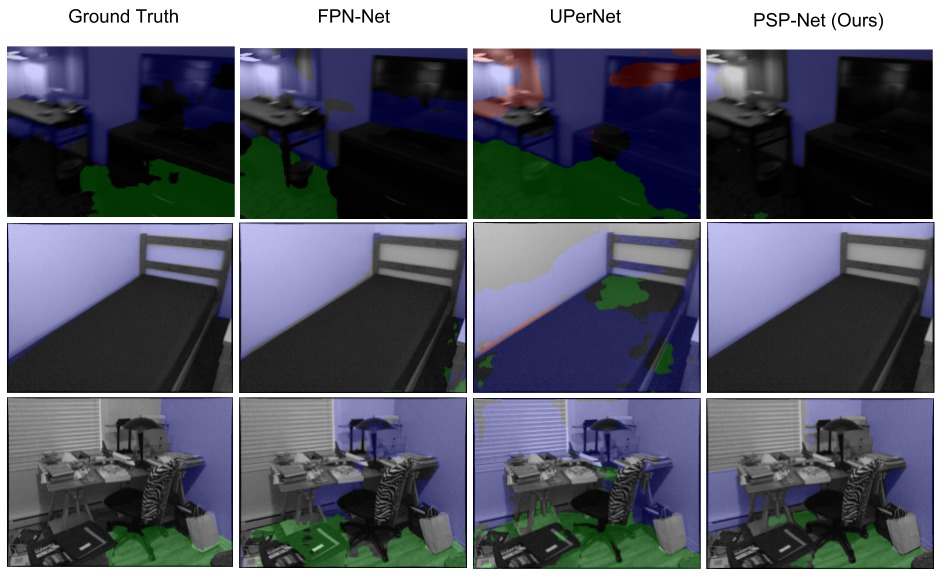}}
\caption{Segmentation Results. Our PSP-Net (our final model) are in the right column. Walls are colored blue while floors are colored green. Note that the ground truth sometimes has many missing values and patchy areas, but qualitatively our results are more robust and consistent.}
    \label{fig:segresults}
\end{figure}

Results for our segmentation submodule are displayed in Table \ref{table:cfwSegmentation}, where we compare to a state-of-the-art model trained on ADE20K and an FPN-Net \cite{lin2017feature} architecture we trained from scratch. We see that our model (with a pyramid scene parsing network architecture \cite{zhao2017pyramid}) performs the best on all metrics. A forward pass through this particular network is efficient and for an image of size $640\times480$ takes $~100ms$. \footnote{Note that these metrics are unreliable due to corrupted ScanNet ground truth (see Fig. \ref{fig:segresults}). Such observations justify our decision not to fine-tune our models on the ScanNet dataset. }



\begin{table}[h]
\centering
\begin{tabular}{|c|c|c|c|c|}
    \hline
    Method &  Wall IoU & Wall F1\\
    \hline
    UPerNet \cite{zhou2018semantic} & 0.406 & 0.512\\
    \hline
    ResNet-101 FPN \cite{lin2017feature} & 0.471 & 0.576\\
    \hline
    Our PSP Network & \textbf{0.484} & \textbf{0.579}\\
    \hline
\end{tabular}
 \vspace{5pt}
\caption {Evaluation of our wall segmentation module on the ScanNet test scenes. Higher is better for all metrics. We present the IoU (True positives over total positives) and F1 (harmonic mean of precision and recall) scores. For both metrics higher is better.}
\label{table:cfwSegmentation}
\end{table}

\subsection{Public Datasets}
\label{sect:pub_data_a}

\subsubsection{ScanNet}
We use the ScanNet dataset \cite{dai2017scannet} as our main evaluation dataset for perimeter metrics. For indoor layout or perimeter estimation from a sequence of RGB images with pose, no other large-scale dataset exists to the best of our knowledge. 
We choose our test set as the scenes with most complete boundaries and coverage, especially on corners, so that ground truth perimeter annotations are possbile. To generate the groundtruth perimeters for evaluation metrics, we hand annotate 76 distinct ScanNet test scenes by marking corners on the provided ScanNet reconstruction in Meshlab \cite{cignoni2008meshlab}. Of the 1513 scenes available, multiple scenes are re-captures of the same environment with different trajectories or minor scene changes. After eliminating the scenes that are duplicates of our chosen test scenes, the remaining 1369 scenes are set aside for training the multi-view depth estimation network.

\subsubsection{ADE20k/SunRGBD}\label{sect:wallsegdata} 

We trained segmentation from scratch on the popular public datasets ADE20K \cite{zhou2018semantic} and then fine-tuned on SunRGBD \cite{song2015sun}. We perform no additional fine-tuning on the target dataset because we found the ScanNet dataset floor/wall/ceiling annotations to be rather incomplete and erroneous (see ground truth examples in Fig. \ref{fig:segresults})

We used the ADE20k dataset in its standard segmentation setting with 150 classes, but the SunRGBD dataset was preprocessed for a 4 way segmentation (floor, wall, ceiling, and other) task. The original SUNRGBD dataset has a 13 class and an extended 37 class labeling scheme for the task of semantic segmentation, which were remapped into our desired classes. We trained on the standard train+val split of SUNRGBD and resized the images to 640 x 480. During training, we used data augmentations like horizontal flips, crop, scaling, contrast, gaussian noise and gaussian blur, including random sequences of these operations. In Sect. \ref{sect:segresults} we see that fine-tuning on this modified wall segmentation variant of SunRGBD gives us a substantial boost in performance on wall segmentation. 



\subsubsection{FloorNet}
In order to test the flexibility and generality of our framework with respect to variations in input modalities and scene types, we test our method on the FloorNet \cite{liu2018floornet} dataset, which includes much larger, multi-room scene types than those in ScanNet. Since the input trajectories were not available for this dataset, we evaluate the clustering and perimeter estimation modules only, starting from the captured FloorNet meshes. This inference is a strictly easier problem than the one our full pipeline solves, but by displaying an excellent performance on this dataset, we are able to show that our pipeline works well for arbitrary layout shapes of various scales and complex room configurations.


\end{document}